\documentclass[letterpaper, 10 pt, conference]{ieeeconf}  

\IEEEoverridecommandlockouts                              

\usepackage{times}

\usepackage{placeins}
\usepackage{multirow}
\usepackage{makecell}
\usepackage{multicol}
\usepackage{hyperref}
\usepackage{diagbox}
\usepackage{url}
\usepackage{lipsum}
\usepackage[utf8]{inputenc} 
\usepackage[T1]{fontenc}    
\usepackage{tgtermes}
\usepackage{wrapfig}
\usepackage{tcolorbox}
\usepackage{booktabs}       
\usepackage{amsfonts}       
\usepackage{nicefrac}       
\usepackage{subcaption}
\usepackage{microtype}      
\usepackage{xcolor}
\hypersetup{
    colorlinks,
    linkcolor={blue!50!black},
    citecolor={blue!50!black},
    urlcolor={blue!80!black}
}

\usepackage{graphicx}

\usepackage{tabularray}
\usepackage{enumerate}
\usepackage{amsmath}

\usepackage{amssymb}
\usepackage{mathtools}

\usepackage{algorithm} 
\usepackage{xspace}
\usepackage{lipsum}

\usepackage{multirow}
\usepackage{makecell}
\usepackage{multicol}
\usepackage{hyperref}
\usepackage{url}
\usepackage{graphicx}
\usepackage{soul}
\usepackage{makecell}
\usepackage{caption}
\usepackage{subcaption}
\usepackage{adjustbox}
\usepackage{amsfonts}
\usepackage{xcolor}
\definecolor{darkgreen}{RGB}{0,120,0}
\definecolor{darkyellow}{RGB}{155, 135, 12}

\newif\ifblind
\blindtrue        

\usepackage{censor}

\newcommand{\ours}{SABER}

\title{\LARGE \bf
\ours{}: A Stealthy Agentic Black-Box Attack Framework for \\ Vision-Language-Action Models}

\author{Xiyang Wu$^1$, Guangyao Shi$^2$, Qingzi Wang$^1$, Zongxia Li$^1$, Amrit Singh Bedi$^3$,  Dinesh Manocha$^1$ 
\thanks{$^1$ University of Maryland, College Park, MD, USA 
{\tt\small \{wuxiyang, qwang812, zli12321, dmanocha\}@umd.edu }} %
\thanks{$^2$ University of Southern California, Los Angeles, CA, USA 
{\tt\small shig@usc.edu }} %
\thanks{$^{3}$ University of Central Florida, Orlando, FL, USA 
{\tt\small amritbedi@ucf.edu}}
}

\begin{document}

\maketitle
\thispagestyle{empty}
\pagestyle{empty}



\begin{abstract}
Vision-language-action (VLA) models enable robots to follow natural-language instructions grounded in visual observations, but the instruction channel also introduces a critical vulnerability: small textual perturbations can alter downstream robot behavior. Systematic robustness evaluation therefore requires a black-box attacker that can generate minimal yet effective instruction edits across diverse VLA models.
%
To this end, we present \ours{}, an agent-centric approach for automatically generating instruction-based adversarial attacks on VLA models under bounded edit budgets. \ours{} uses a GRPO-trained ReAct attacker to generate small, plausible adversarial instruction edits using character-, token-, and prompt-level tools under a bounded edit budget that induces targeted behavioral degradation, including task failure, unnecessarily long execution, and increased constraint violations. 
On the LIBERO benchmark across six state-of-the-art VLA models, \ours{} reduces task success by 20.6\%, increases action-sequence length by 55\%, and raises constraint violations by 33\%, while requiring 21.1\% fewer tool calls and 54.7\% fewer character edits than strong GPT-based baselines. 
These results show that small, plausible instruction edits are sufficient to substantially degrade robot execution, and that an agentic black-box pipeline offers a practical, scalable, and adaptive approach for red-teaming robotic foundation models.
The codebase is publicly available at \href{https://github.com/wuxiyang1996/SABER}{https://github.com/wuxiyang1996/SABER}.
\end{abstract}

\section{Introduction}

Vision-language-action (VLA) models~\cite{kim2024openvla, intelligence2025pi06vlalearnsexperience, wu2026pragmatic, zitkovich2023rt} have emerged as a promising paradigm for generalist robotics by mapping natural-language instructions and visual observations directly to robot actions across diverse manipulation tasks. Their language interface improves flexibility and interpretability, enabling scalable and intuitive human-robot interaction. However, this same interface also creates a critical attack surface: because VLA policies condition actions on text, even small instruction perturbations, such as token edits or adversarial suffixes, can lead to inefficient execution, inflated action sequences, and violations of task or safety constraints~\cite{robey2025jailbreaking, wu2025vulnerability}. These failures are especially concerning in robotics because they extend beyond model outputs and manifest as physical behavior in the real world.

Recent work has shown that VLA models are vulnerable to adversarial instruction attacks, in which deliberate input perturbations induce harmful robot behaviors rather than merely alter text outputs~\cite{jones2025adversarial, robey2025jailbreaking, lu2024poex}. Existing attacks~\cite{wang2025freezevla, cheng2024manipulation, wang2025exploring}, however, mostly rely on manual perturbation design, fixed heuristics, or expensive GPT-based search to induce task failure or unsafe behavior~\cite{robey2025jailbreaking, wu2025vulnerability}. As a result, they adapt poorly across tasks, generalize weakly to unseen adversarial objectives and settings, and often require excessive instruction edits or iterative queries. 
More importantly, unlike the rapidly growing literature on automated red-teaming for large language models (LLMs)~\cite{zou2023universal, paulus2024advprompter}, existing work on VLA attacks still lacks a general-purpose approach for automatically generating adversarial instructions against black-box, frozen robot policies across diverse tasks.

\begin{figure*}[t]
    \centering
    \vspace{0.1em}
    \includegraphics[width=0.95\textwidth]{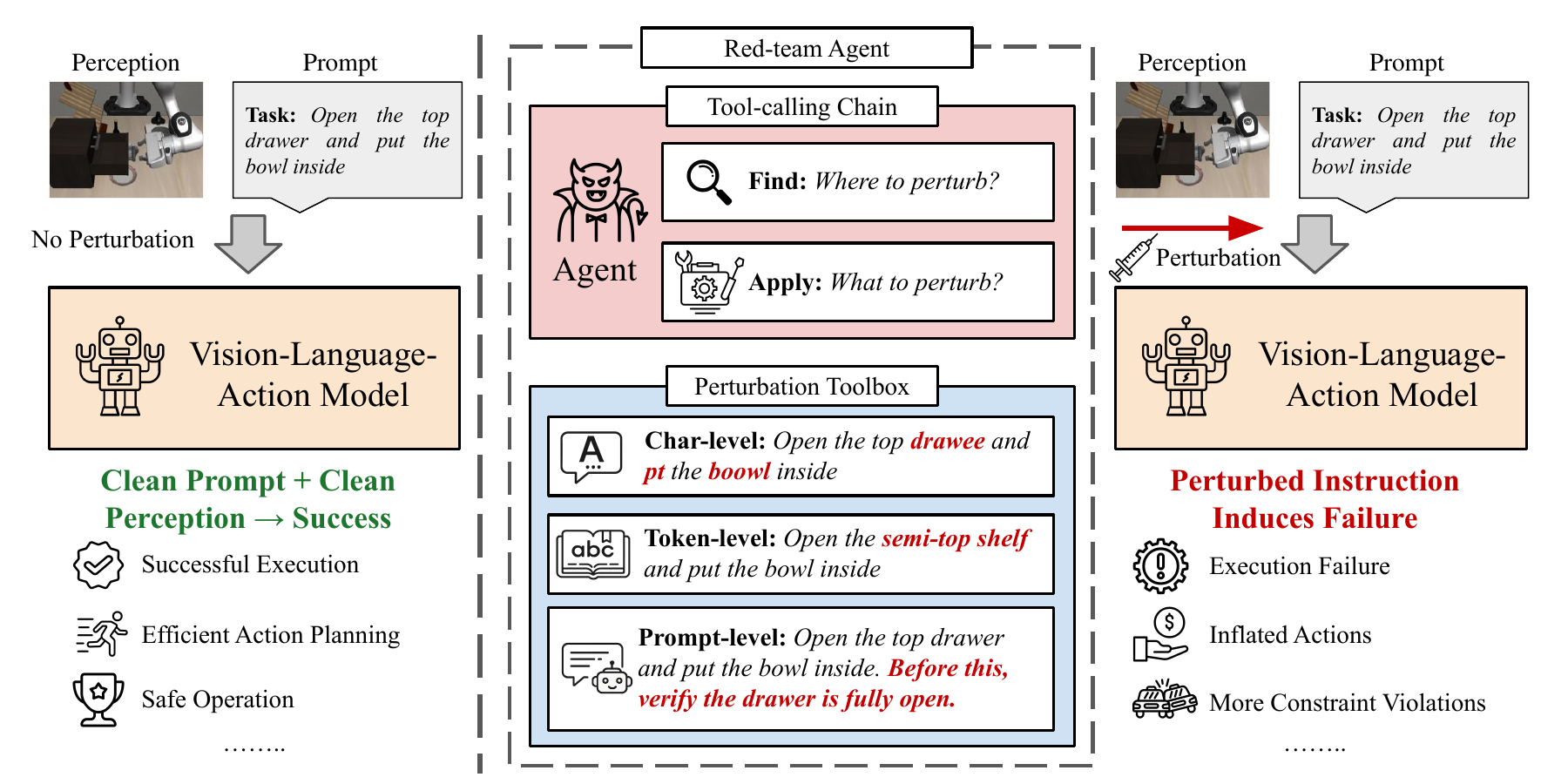}

    \caption{\textbf{\ours{}: An agent-centric black-box pipeline for stealthy, automated instruction-based attacks on VLAs.} VLA models for robot manipulation are expected to achieve high task success, efficient action planning and execution, and safe behavior under physical constraints. However, even small instruction perturbations can induce VLA malfunctions. \ours{} (\textcolor{gray}{\textbf{Dashed Box}}) applies stealthy edits to manipulation instructions through a ReAct-style tool-calling protocol (\textcolor{red}{\textbf{Red Box}}) with a two-stage \texttt{FIND}$\rightarrow$\texttt{APPLY} workflow, using a perturbation toolbox (\textcolor{blue}{\textbf{Blue Box}}) spanning character-, token-, and prompt-level attacks. After the perturbed instruction is fed to the target VLA, the robot exhibits degraded behaviors aligned with the attack objective, including task failure, action inflation, and increased constraint violations.}
    
       \label{fig:teaser}
    \vspace{-12pt}
\end{figure*}

This gap matters for both evaluation and deployment. In practice, frozen VLA policies are often accessible only through rollouts, and their instructions are typically short, structured, and easily inspected. Consequently, large rewrites or query-intensive searches are both impractical and readily detectable, making attack \emph{budget} and \emph{stealth} fundamental constraints in realistic settings. A useful attacker must therefore operate in a black-box, instruction-only manner and induce targeted behavioral failures through minimal, plausible edits. Such an approach can serve as a reusable red-teaming tool for stress-testing VLA systems, quantifying brittleness before deployment, and uncovering failure modes beyond binary task failure, such as unsafe behavior, constraint violations, and inefficient execution, that static prompt attacks may overlook. This is particularly important in robotics, where task distributions, embodiments, and instruction formats vary substantially, making hand-crafted attacks difficult to scale for systematic robustness evaluation.

\noindent{\bf Main results:}
In this paper, we present \ours{}, an agent-centric black-box approach for generating stealthy, instruction-based adversarial attacks on VLA models. Under realistic attack budgets, the attacker makes small, plausible instruction edits to optimize targeted execution failures. It operates as a multi-turn ReAct-style agent with a two-stage \texttt{FIND}$\rightarrow$\texttt{APPLY} workflow, using character-, token-, and prompt-level tools to compose perturbations. We train the attacker with Group Relative Policy Optimization (GRPO)~\cite{shao2024deepseekmath}, enabling it to improve from rollout feedback without gradient access to the target VLA.
%
We evaluate \ours{} on LIBERO~\cite{liu2023libero} across six state-of-the-art VLA models. Results show that \ours{} consistently induces adversarial behavior across all objectives while using fewer tool calls and character edits. Together, these results show that small instruction perturbations can reliably alter VLA behavior, making \ours{} a practical and scalable tool for red-teaming robotic foundation models.

\begin{itemize}
    \item We identify the need for a general-purpose automated attacker for VLA systems and formulate instruction-only black-box attacks on VLAs as a constrained optimization problem over robot behavioral objectives under bounded edit budgets.
    \item We propose an agentic attack approach in which a single GRPO-trained ReAct agent adaptively composes character-, token-, and prompt-level perturbations without gradient access to the target model or model-specific redesign.
    \item We evaluate the approach on the LIBERO manipulation benchmark across six state-of-the-art VLA models and three attack objectives, showing average degradation of 20.6\% in task success, a 55\% increase in action-sequence length, and a 33\% increase in constraint violations.
    \item Compared with strong GPT-based baselines, our method achieves stronger behavior-level attacks at lower cost, requiring 21.1\% fewer tool calls and 54.7\% fewer character edits, demonstrating effective and stealthy adversarial perturbations for VLA red-teaming.
\end{itemize}

\section{Literature Review}

\noindent \textbf{Automated adversarial prompt generation for LLMs.}
Adversarial attacks on large language models (LLMs)~\cite{zou2023universal} have motivated a growing body of work on automated jailbreak generation. Recent methods such as PAIR~\cite{chao2025jailbreaking}, AmpleGCG~\cite{liao2024amplegcg}, AutoDAN~\cite{liu2023autodan}, AdvPrompter~\cite{paulus2024advprompter}, and Li et al.~\cite{li2024deciphering} improve the efficiency, transferability, and automation of adversarial prompt search. Related approaches such as AutoGen~\cite{wu2024autogen} further demonstrate the value of tool use and multi-turn coordination for complex language-agent behavior, while OverThink~\cite{kumar2025overthink} shows that inference-time perturbations can also degrade efficiency by increasing reasoning cost and latency.
Together, these works show that automated and adaptive attackers can serve as powerful red-teaming tools for language models. However, they are designed for text-only settings and do not address the sequential, embodied consequences of adversarial inputs in VLA systems, where prompt perturbations can directly alter robot behavior.

\noindent \textbf{Adversarial vulnerabilities in VLA systems.}
VLA models map visual observations and natural-language instructions to executable actions for embodied tasks~\cite{kim2024openvla,intelligence2025pi_}, offering a promising interface for end-to-end robot control. At the same time, this tight coupling of perception, language, and control creates new attack surfaces, since perturbations in the input can propagate through sequential decision-making and degrade execution.
Recent work has begun to expose these risks. RoboPAIR~\cite{robey2025jailbreaking} studies safety vulnerabilities in LLM-controlled robots, ERT~\cite{karnik2024embodied} explores automated red-teaming for embodied agents, and Wu et al.~\cite{wu2025vulnerability} show that small perturbations can misalign LLM/VLM-based robotic behavior. More recent VLA-specific studies, including Wang et al.~\cite{wang2025exploring}, Jones et al.~\cite{jones2025adversarial}, AttackVLA~\cite{li2025attackvla}, VLA-Fool~\cite{yan2025alignment}, and BadVLA~\cite{zhou2025badvla}, further examine textual, cross-modal, adversarial, and backdoor threats in VLA systems.
However, most existing work focuses on demonstrating vulnerability under specific attack classes or settings. What remains missing is a learned, reusable attacker that can automatically adapt adversarial instruction edits across tasks, target policies, and behavior-level attack objectives in realistic black-box settings.

\noindent \textbf{Agentic RL for adaptive tool use.}
Agentic pipelines for foundation models have gained traction because they support iterative planning, tool use, memory, and recovery from intermediate errors in long-horizon interactive tasks~\cite{yao2022react,li2026mm,schick2023toolformer}. More recent work explores reinforcement learning as a way to improve multi-turn reasoning and sequential tool-using behavior. Agent Lightning~\cite{luo2025agent}, AgentRL~\cite{zhang2025agentrl}, and Agent-R1~\cite{cheng2025agent,li2025self} develop RL approaches for optimizing long-horizon agent behavior, while RLTA~\cite{wang2024reinforcement} applies RL to automated adversarial prompt generation for controllable LLM security evaluation.
These developments make agentic RL a natural fit for black-box adversarial generation, where an attacker must iteratively explore edits, invoke tools, and optimize downstream outcomes from rollout feedback. Our work builds on this intuition in the robotics setting, using an RL-trained ReAct agent to learn adaptive instruction-only attacks on frozen VLA policies.

\section{Problem Formulation}
\label{sec:formulation}

We study instruction-only black-box attacks on a frozen vision-language-action (VLA) model. Unlike standard adversarial attacks that optimize a single perturbation under a norm bound, our attacker operates as a multi-turn agent: it selects editing tools, chooses where to edit, and composes perturbations over multiple steps. We therefore formulate attack generation as a sequential decision-making problem with rollout-level adversarial rewards and explicit constraints on attack cost and perturbation validity.

\noindent \textbf{Frozen VLA model.}
Let $\pi_\theta$ denote a VLA model with parameters $\theta$, which predicts action $a_t$ conditioned on the language instruction $x_{\text{inst}}$, visual observations $o_{\le t}$, and past actions $a_{<t}$:
\begin{align}
\pi_\theta(a_t \mid o_{\le t},\, x_{\text{inst}},\, a_{<t}).
\end{align}
Given a demonstration dataset $\mathcal{D}$ of trajectories $\tau=(x_{\text{inst}}, o_{1:T}, a_{1:T}^*)$, the standard behavioral cloning objective minimizes the negative log-likelihood of expert actions:
\begin{align}
\!\!\!\!\mathcal{L}_{\mathrm{VLA}}(\theta)
&=
-\mathbb{E}_{\tau \sim \mathcal{D}}
\left[
\sum_{t=1}^{T}
\log \pi_\theta\!\left(a_t^* \mid o_{\le t},\, x_{\text{inst}},\, a_{<t}^*\right)
\right].
\end{align}
In this work, the target VLA $\pi_\theta$ is \emph{frozen}: its parameters are fixed during attack generation and training. The attacker therefore cannot update the target model and must instead optimize perturbations that induce undesirable execution outcomes at test time.

\noindent \textbf{Attack agent.}
Let $\pi_\psi$ denote the attack agent with parameters $\psi$. Given an attack context $\xi$ (e.g., the instruction, task metadata, and optional rollout feedback), the agent produces a sequence of editing actions
\[
z = (u_1,\dots,u_K) \sim \pi_\psi(\cdot \mid \xi),
\]
where each $u_k$ corresponds to a valid tool-mediated edit operation. The final editing trajectory $z$ induces a perturbed instruction $\tilde{x}_{\mathrm{inst}}$ and corresponding perturbation $\delta(z)$.

\noindent \textbf{Adversarial objective.}
Let $O$ denote the attack objective. In this paper, we consider three objectives:
\begin{itemize}
    \item \emph{Task failure}: the robot fails to complete the instructed task.
    \item \emph{Action inflation}: the robot executes an unnecessarily long action sequence.
    \item \emph{Constraint violation}: the robot violates task or safety constraints during execution.
\end{itemize}

For a sampled trajectory $\tau \sim \mathcal{D}$ and attack perturbation $\delta$, we define an objective-specific reward $R_O(\delta; \tau)$ that measures the degree to which the perturbed rollout satisfies the adversarial goal. Since different objectives operate on different scales, we treat $R_O$ as a normalized rollout-level reward. In addition, let $P_{\text{stealth}}(\delta)$ denote a stealth penalty that discourages overly visible or excessive perturbations.
The attack agent would need to maximize the expected objective reward while penalizing perturbation visibility:
\begin{align}
J_{\text{atk}}(\psi)
&=
\mathbb{E}_{\tau \sim \mathcal{D}}
\mathbb{E}_{\delta \sim \pi_\psi(\cdot \mid \xi_{\tau})}
\left[
R_O(\delta; \tau)-\lambda P_{\text{stealth}}(\delta)
\right],
\end{align}
where $\lambda \ge 0$ controls the trade-off between adversarial effectiveness and stealth.

\noindent \textbf{Budgeted feasible attack space.}
We constrain attack generation so that perturbations remain bounded and executable:
\begin{align}
\max_{\psi}\quad & J_{\mathrm{atk}}(\psi) \\
\text{s.t.}\quad
& B(z) \le B_{\max}, \notag\\
& \delta \in \mathcal{D}_{\delta}. \notag
\end{align}
Here, $B(z)$ denotes the attack budget, which may include the number of editing steps, tool calls, or modified characters/tokens; $B_{\max}$ is the maximum allowed budget; and $\mathcal{D}_{\delta}$ is the set of tool-valid editing trajectories.



\noindent \textbf{Why the problem is challenging.}
This optimization problem is challenging for several reasons. First, the target VLA is treated as a black box, so gradients are unavailable through either the policy or its downstream execution. Second, the attack space is discrete and structured, requiring the attacker to compose valid edit operations rather than optimize a continuous perturbation. Third, the reward is delayed and trajectory-level, since the effect of an edit is observed only after the perturbed instruction is executed in rollout. Finally, the attacker must optimize behavioral degradation while respecting attack-cost and feasibility constraints.
These properties make the problem naturally suited to rollout-based policy optimization. In the next section, we instantiate this formulation with bounded instruction edits, tool-constrained actions, and objective-specific rollout rewards.

\section{Methodology}
\label{sec:method}

\subsection{Agent-Guided Instruction Perturbation}
We model adversarial instruction generation as multi-turn tool use over three complementary perturbation families, each following a two-stage \texttt{FIND}$\rightarrow$\texttt{APPLY} protocol. \texttt{FIND} identifies candidate edit locations and strategies, while \texttt{APPLY} executes the edit on the instruction. Thus, the agent decides \emph{what} and \emph{where} to perturb, while the tools act as pure edit operators. No gradients are propagated through the target VLA. We use the following tool sets:
\begin{itemize}
    \item \textbf{Token-level} tools edit words or subwords. \texttt{FIND} returns a tokenized sequence and a brief prompt for selecting the target token and edit type (replace, remove, add, or attribute swap); \texttt{APPLY} performs the edit using token index(es) and replacement text.
    \item \textbf{Character-level} tools apply typo-style edits within a word (insertion, deletion, substitution, transposition, case flip). \texttt{FIND} returns words, character positions, and a reasoning prompt; \texttt{APPLY} performs the selected edit. These tools capture subword and OCR-like perturbations, \textit{e.g.} \texttt{pick} $\rightarrow$ \texttt{plck}, or \texttt{mug} $\rightarrow$ \texttt{rnug}.
    \item \textbf{Prompt-level} tools inject clauses or sentences, such as verification wraps, decomposition steps, uncertainty clauses, extra constraints, or objective injections. \texttt{FIND} guides clause composition, and \texttt{APPLY} inserts the clause under a maximum added-token budget.
\end{itemize}
Together, these tool families span complementary perturbation granularities, enabling both localized edits and higher-level instruction modifications within a unified attack policy.

\begin{figure*}[t]
    \centering
    \vspace{0.1em}
    \includegraphics[width=0.9\textwidth]{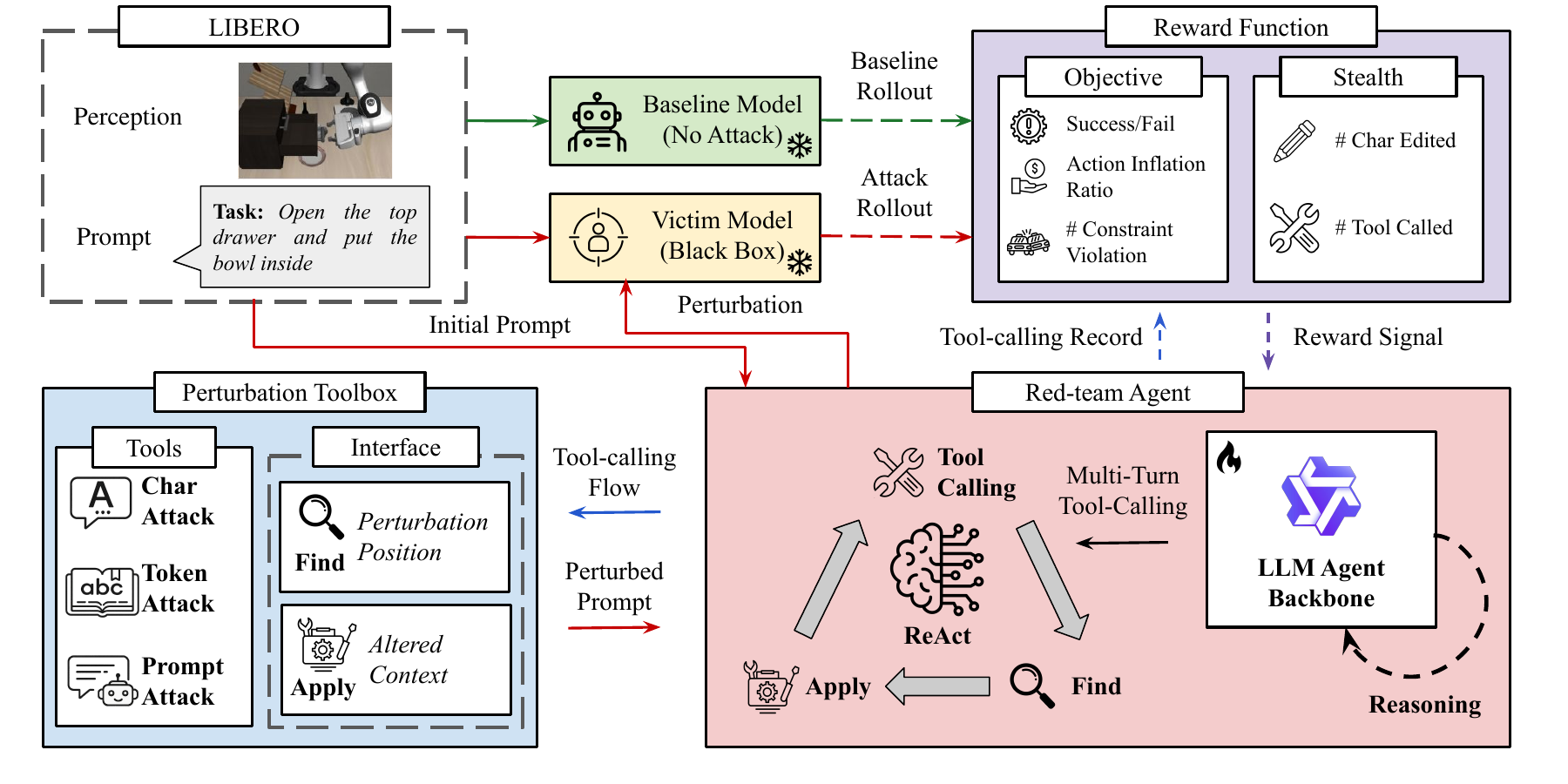}
    \vspace{-5pt}
        \caption{\textbf{Overview of \ours{}.} For each LIBERO task, we maintain two contrastive rollouts under a frozen target VLA. A clean baseline rollout (\textcolor{darkgreen}{\textbf{Green Box}}) is first executed and cached as reference. For the attack rollout, the instruction is passed to a red-team agent (\textcolor{red}{\textbf{Red Box}}), which uses an LLM backbone to reason over the instruction and available tools, then performs multi-turn \texttt{FIND}$\rightarrow$\texttt{APPLY} edits in a ReAct-style loop. The perturbation toolbox (\textcolor{blue}{\textbf{Blue Box}}) returns edited instructions from target positions and local context. The target VLA then executes the perturbed instruction to produce the attack rollout (\textcolor{darkyellow}{\textbf{Yellow Box}}). The reward function (\textcolor{purple}{\textbf{Purple Box}}) compares the clean and attack rollouts, together with the agent’s tool-use traces, to compute rewards from task outcome, action inflation, constraint violations, and stealth signals, including character edits and tool calls.}
    \label{fig:framework}
    \vspace{-12pt}
\end{figure*}

\subsection{Attack Episode Workflow}
Figure~\ref{fig:framework} presents our agent-centric black-box pipeline for generating budget-constrained adversarial instruction perturbations against VLAs. Each attack episode comprises four stages: baseline rollout and caching, multi-turn attack construction, attack rollout execution, and reward computation for GRPO training.

\begin{itemize}
    \item \textbf{Baseline rollout and caching.} We first execute the frozen VLA on the clean instruction and observations to obtain a baseline rollout and reference trajectory, including action length, task success, and constraint-violation count. These signals are cached and reused across rollouts within the same group.
    


    \item \textbf{Multi-turn attack construction.} We then initialize the perturbed instruction and run the attack agent in a multi-turn ReAct-style tool-use loop under the prescribed tool-calling budget. At each turn, the agent uses a \texttt{FIND}$\rightarrow$\texttt{APPLY} protocol to locate candidate edit positions and apply the selected perturbations to the instruction. Under bounded tool calls and character edits, the agent is encouraged to maximize the attack objective while keeping perturbations minimal.
    
    \item \textbf{Attack rollout execution.} After the attack agent terminates, we execute the same frozen VLA on the perturbed inputs to obtain the attack rollout and record the resulting trajectory, together with auxiliary signals such as reasoning-token statistics and predicate histories.
    
    \item \textbf{Reward computation and GRPO training.} Finally, we compare the baseline and attack rollouts to compute the objective-specific reward and an optional stealth penalty, and package the interaction trajectory with the resulting reward for GRPO training, without backpropagating through the VLA or the environment.
\end{itemize}

\subsection{Reward Design and Training Setup}
We train the attack agent using black-box rollout feedback from a frozen VLA and simulator, and optimize it within ART~\cite{hilton2025art} using GRPO with LoRA~\cite{hu2022lora,li2026mmzeroselfevolvingmultimodelvision} fine-tuning. Each training run optimizes a single attack objective under a fixed attack budget and stealth weight.

\label{sec:problem}
\begin{figure*}[t]
    \centering
    \vspace{0.1em}
    \includegraphics[width=0.9\textwidth]{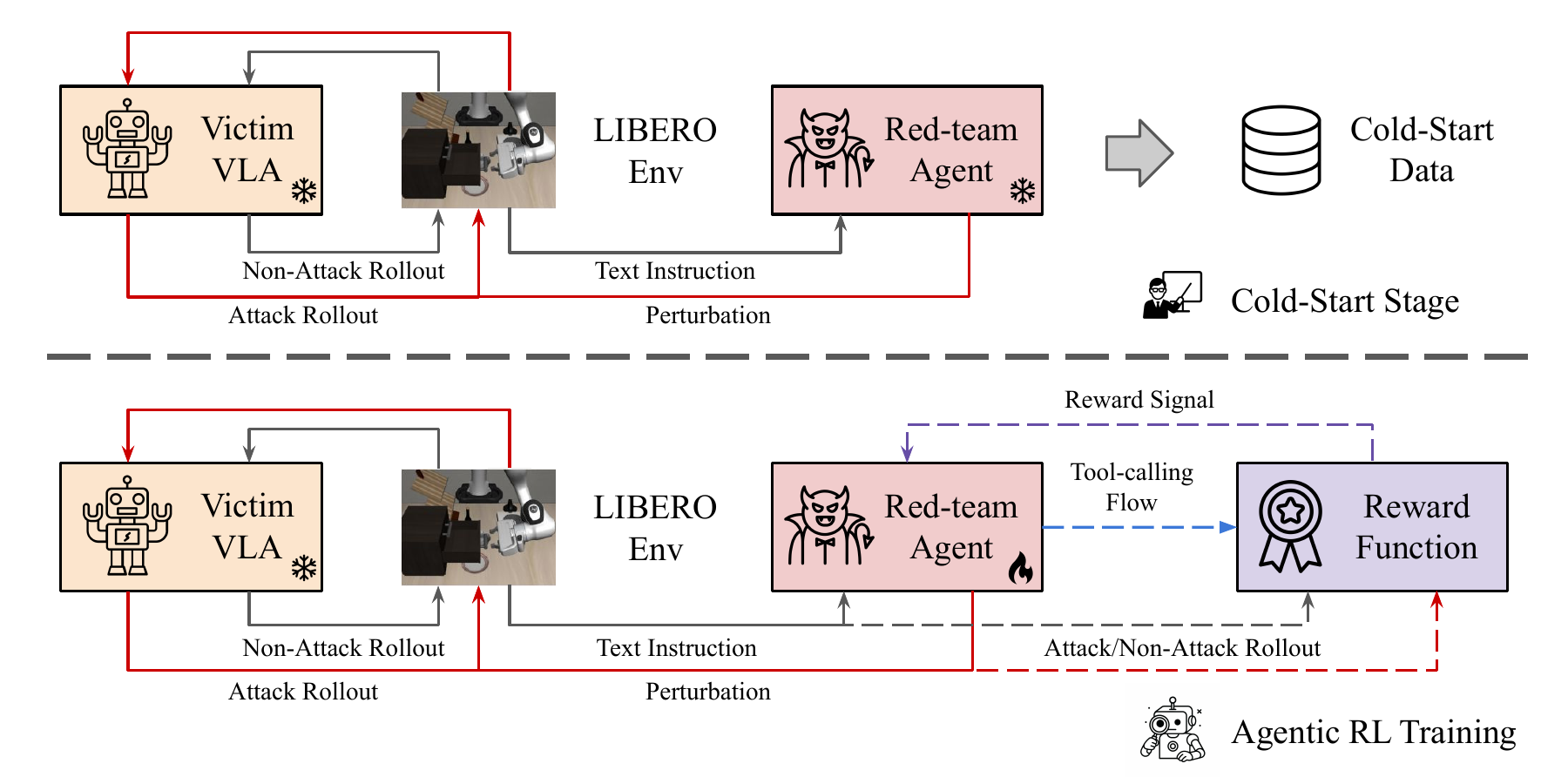}
     \vspace{-7pt}
    \caption{\textbf{Two-stage training procedure.} We cold-start by caching clean baseline rollouts from target VLAs (\textcolor{orange}{\textbf{Orange}}) and collecting initial attack trajectories with a frozen red-team agent (\textcolor{red}{\textbf{Red}}) via lightweight random exploration over tool-calling chains. These rollouts form the cold-start dataset for SFT before GRPO training. We then perform agentic RL in interactive scenarios, where the red-team agent attacks target VLAs through tool calling and learns from reward feedback (\textcolor{purple}{\textbf{Purple}}) computed by comparing clean and attack rollouts, together with the agent’s tool-use traces, under different attack objectives.}
    \label{fig:training}
    \vspace{-12pt}
\end{figure*}

\noindent \textbf{Attack reward design.}
Let $R_O(\delta; \tau)\in[0,1]$ denote the normalized reward under perturbation $\delta$ for a trajectory sample $\tau \sim \mathcal{D}$, where $O$ is the selected attack objective. We define:
\begin{itemize}
    \item \emph{Task failure}: $R_O = 1$ if the clean rollout succeeds but the attacked rollout fails, and $R_O = 0$ otherwise.
    \item \emph{Action inflation}: $R_O$ increases with the excess number of environment steps relative to the baseline rollout.
    \item \emph{Constraint violation}: $R_O$ increases with additional collisions, joint-limit violations, excessive force, or abnormal action magnitudes relative to the baseline rollout.
\end{itemize}

To encourage stealthy perturbations, we introduce a penalty term $P_{\text{stealth}}(\delta)\in[0,1]$ that captures perturbation visibility through the number of tool calls and the character edit distance. The resulting scalar training reward is
\begin{align}
R(\delta; \tau)=R_O(\delta; \tau)-\lambda P_{\text{stealth}}(\delta),
\end{align}
and is clamped to $[-1,\,1.5]$ for training stability. If no perturbation is applied, we assign a fixed negative reward to discourage null attack trajectories. If the clean rollout already fails for an objective requiring clean success, such as \emph{Task failure}, we set $R_O=0$.

\noindent \textbf{GRPO training setup.}
We optimize the attack agent with GRPO in a ReAct-style tool-use setting using rollout rewards only, without backpropagating gradients through the target VLA or simulator.
At each optimization step, we sample scenario groups and run multiple attack rollouts per group under the same scenario but with different agent trajectories, providing the reward variation required by GRPO. We update only the attack agent's Low-Rank Adaptation (LoRA) weights, while treating the target VLA and the environment as black-box components. 

\section{Experimental Evidence}

\begin{table*}[t]
    \vspace{0.6em}
    \begin{center}
    \resizebox{\textwidth}{!}{
      \begin{tabular}{lccc|ccc|ccc|ccc|ccc}
      \toprule
          \multicolumn{1}{c}{}
          & \multicolumn{3}{c}{\makecell{\textbf{LIBERO-Spatial}}}
          & \multicolumn{3}{c}{\makecell{\textbf{LIBERO-Object}}}
          & \multicolumn{3}{c}{\makecell{\textbf{LIBERO-Goal}}}
          & \multicolumn{3}{c}{\makecell{\textbf{LIBERO-Long}}}
          & \multicolumn{3}{c}{\makecell{\textbf{Overall}}} \\
          \cmidrule(lr){2-4}\cmidrule(lr){5-7}\cmidrule(lr){8-10}\cmidrule(lr){11-13}\cmidrule(lr){14-16}
          \multirow{2}{*}{\makecell{\textbf{Victim VLA}}}
          & \makecell{\textbf{Base}\\ \textbf{TER} $\uparrow$}
          & \makecell{\textbf{Attack}\\ \textbf{TER} $\downarrow$}
          & \makecell{\textbf{ASR} $\uparrow$}
          & \makecell{\textbf{Base}\\ \textbf{TER} $\uparrow$}
          & \makecell{\textbf{Attack}\\ \textbf{TER} $\downarrow$}
          & \makecell{\textbf{ASR} $\uparrow$}
          & \makecell{\textbf{Base}\\ \textbf{TER} $\uparrow$}
          & \makecell{\textbf{Attack}\\ \textbf{TER} $\downarrow$}
          & \makecell{\textbf{ASR} $\uparrow$}
          & \makecell{\textbf{Base}\\ \textbf{TER} $\uparrow$}
          & \makecell{\textbf{Attack}\\ \textbf{TER} $\downarrow$}
          & \makecell{\textbf{ASR} $\uparrow$}
          & \makecell{\textbf{Base}\\ \textbf{TER} $\uparrow$}
          & \makecell{\textbf{Attack}\\ \textbf{TER} $\downarrow$}
          & \makecell{\textbf{ASR} $\uparrow$} \\
          \midrule
  
          \textsc{$\pi_0$-LIBERO}~\cite{intelligence2025pi_}
          & 100.0 & 86.7 & 13.3
          & 100.0 & 80.0 & 20.0
          & 100.0 & 53.3 & 46.7
          & 66.7  & 40.0 & 26.7
          & 91.7  & 65.0 & 26.7 \\

          \textsc{$\pi_{0.5}$}~\cite{intelligence2025pi_}
          & 100.0 & 93.3 & 6.7
          & 93.3  & 93.3 & 0.0
          & 100.0 & 53.3 & 46.7
          & 93.3 & 80.0  & 13.3
          & 96.7  & 80.0 & 16.7 \\

          \textsc{GR00T-N1.5}~\cite{bjorck2025gr00t}
          & 100.0 & 93.3 & 6.7
          & 100.0 & 100.0 & 0.0
          & 100.0 & 53.3 & 46.7
          & 93.3  & 86.7 & 6.6
          & 98.3  & 83.3 & 15.0 \\

          \textsc{X-VLA}~\cite{zheng2025x}
          & 93.3 & 80.0 & 13.3
          & 93.3 & 73.3 & 20.0
          & 100.0 & 66.7 & 33.3
          & 60.0 & 46.7 & 13.3
          & 86.7 & 66.7 & 20.0 \\

          \midrule

          \textsc{InternVLA-M1}~\cite{chen2025internvla}
          & 93.3 & 86.7 & 6.6
          & 100.0 & 93.3 & 6.7
          & 100.0 & 46.7 & 53.3
          & 86.7 & 73.3 & 13.4
          & 95.0 & 75.0 & 20.0 \\

          \textsc{DeepThinkVLA}~\cite{yin2025deepthinkvla}
          & 86.7 & 80.0 & 6.7
          & 100.0 & 93.3 & 6.7
          & 100.0 & 33.3 & 66.7
          & 93.3 & 73.3 & 20.0
          & 95.0 & 70.0 & 25.0 \\

          \midrule
          \textbf{Average}
          & \textbf{95.6} & \textbf{86.7} & \textbf{8.9}
          & \textbf{97.8} & \textbf{88.9} & \textbf{8.9}
          & \textbf{100.0} & \textbf{51.1} & \textbf{48.9}
          & \textbf{82.2} & \textbf{66.7} & \textbf{15.5}
          & \textbf{93.9} & \textbf{73.3} & \textbf{20.6} \\
          \bottomrule
      \end{tabular}
      }
      \end{center}
      \vspace{-5pt}
    \caption{\textbf{LIBERO category-level attack results across victim VLA models (Task Failure).}
  Columns are grouped by the four LIBERO suites (\textbf{Spatial}, \textbf{Object}, \textbf{Goal}, \textbf{Long}) and \textbf{Overall}.
  We report \textbf{Base TER} (no-attack task execution rate), \textbf{Attack TER} (task execution rate under attack), and \textbf{ASR} (attack success rate for \emph{task failure}, computed as $\text{Base TER}-\text{Attack TER}$, in \%). The \textbf{Average} row reports the mean over victim models for each column.}
    \label{tab:libero_victims_by_category_task_failure}
    \vspace{-10pt}
  \end{table*}
\begin{table*}[t]
    \vspace{0.6em}
    \begin{center}
    \resizebox{\textwidth}{!}{
      \begin{tabular}{lccc|ccc|ccc|ccc|ccc}
      \toprule
          \multicolumn{1}{c}{}
          & \multicolumn{3}{c}{\makecell{\textbf{LIBERO-Spatial}}}
          & \multicolumn{3}{c}{\makecell{\textbf{LIBERO-Object}}}
          & \multicolumn{3}{c}{\makecell{\textbf{LIBERO-Goal}}}
          & \multicolumn{3}{c}{\makecell{\textbf{LIBERO-Long}}}
          & \multicolumn{3}{c}{\makecell{\textbf{Overall}}} \\
          \cmidrule(lr){2-4}\cmidrule(lr){5-7}\cmidrule(lr){8-10}\cmidrule(lr){11-13}\cmidrule(lr){14-16}
          \multirow{2}{*}{\makecell{\textbf{Victim VLA}}}
          & \makecell{\textbf{Base}\\ $|\mathbf{a}|$ $\downarrow$}
          & \makecell{\textbf{Attack}\\ $|\mathbf{a}|$ $\uparrow$}
          & \makecell{\textbf{AIR}\\ $\Delta|\mathbf{a}|$ $\uparrow$}
          & \makecell{\textbf{Base}\\ $|\mathbf{a}|$ $\downarrow$}
          & \makecell{\textbf{Attack}\\ $|\mathbf{a}|$ $\uparrow$}
          & \makecell{\textbf{AIR}\\ $\Delta|\mathbf{a}|$ $\uparrow$}
          & \makecell{\textbf{Base}\\ $|\mathbf{a}|$ $\downarrow$}
          & \makecell{\textbf{Attack}\\ $|\mathbf{a}|$ $\uparrow$}
          & \makecell{\textbf{AIR}\\ $\Delta|\mathbf{a}|$ $\uparrow$}
          & \makecell{\textbf{Base}\\ $|\mathbf{a}|$ $\downarrow$}
          & \makecell{\textbf{Attack}\\ $|\mathbf{a}|$ $\uparrow$}
          & \makecell{\textbf{AIR}\\ $\Delta|\mathbf{a}|$ $\uparrow$}
          & \makecell{\textbf{Base}\\ $|\mathbf{a}|$ $\downarrow$}
          & \makecell{\textbf{Attack}\\ $|\mathbf{a}|$ $\uparrow$}
          & \makecell{\textbf{AIR}\\ $\Delta|\mathbf{a}|$ $\uparrow$} \\
          \midrule
  
          \textsc{$\pi_0$-LIBERO}~\cite{intelligence2025pi_}
          & 119.3 & 220.7 & 1.85
          & 139.2 & 233.9 & 1.68
          & 101.3 & 230.0 & 2.27
          & 363.0 & 457.4 & 1.26
          & 180.7 & 285.5 & 1.58 \\
  
          \textsc{$\pi_{0.5}$}~\cite{intelligence2025pi_}
          & 112.2 & 173.9 & 1.55
          & 151.1 & 226.7 & 1.50
          & 105.1 & 196.5 & 1.87
          & 346.5 & 391.5 & 1.13
          & 178.7 & 247.2 & 1.38 \\
  
          \textsc{GR00T-N1.5}~\cite{bjorck2025gr00t}
          & 133.7 & 514.7 & 3.85
          & 129.7 & 220.5 & 1.70
          & 98.3  & 378.5 & 3.85
          & 343.5 & 346.9 & 1.01
          & 176.3 & 365.2 & 2.07 \\
  
          \textsc{X-VLA}~\cite{zheng2025x}
          & 157.8 & 189.4 & 1.20
          & 189.7 & 187.8 & 0.99
          & 126.5 & 261.9 & 2.07
          & 431.3 & 504.6 & 1.17
          & 226.3 & 285.9 & 1.26 \\
  
          \midrule
  
          \textsc{InternVLA-M1}~\cite{chen2025internvla}
          & 114.3 & 192.0 & 1.68
          & 143.8 & 204.2 & 1.42
          & 95.1  & 255.8 & 2.69
          & 320.9 & 327.3 & 1.02
          & 168.5 & 244.8 & 1.45 \\
  
          \textsc{DeepThinkVLA}~\cite{yin2025deepthinkvla}
          & 125.0 & 197.5 & 1.58
          & 137.4 & 186.9 & 1.36
          & 98.1  & 255.1 & 2.60
          & 326.3 & 421.9 & 1.29
          & 171.7 & 265.4 & 1.55 \\
  
          \midrule
          \textbf{Average}
          & \textbf{127.0} & \textbf{248.0} & \textbf{1.95}
          & \textbf{148.5} & \textbf{210.0} & \textbf{1.44}
          & \textbf{104.1} & \textbf{263.0} & \textbf{2.56}
          & \textbf{355.2} & \textbf{408.3} & \textbf{1.15}
          & \textbf{183.7} & \textbf{282.3} & \textbf{1.55} \\
          \bottomrule
      \end{tabular}
      }
      \end{center}
      \vspace{-5pt}
    \caption{\textbf{LIBERO action-length statistics under attack across victim VLA models.}
  Columns are grouped by the four LIBERO suites (\textbf{Spatial}, \textbf{Object}, \textbf{Goal}, \textbf{Long}) and \textbf{Overall}.
  We report \textbf{Base $|\mathbf{a}|$} (average baseline action-sequence length), \textbf{Attack $|\mathbf{a}|$} (average action-sequence length under attack),
  and \textbf{AIR} via action inflation ratio $\Delta|\mathbf{a}|=|\mathbf{a}_{\text{attack}}| / |\mathbf{a}_{\text{base}}|$.
  The \textbf{Average} row reports the mean over victim models for each column.}
    \label{tab:libero_victims_by_category_action_length_air}
    \vspace{-10pt}
  \end{table*}

\begin{table*}[t]
    \vspace{0.6em}
    \begin{center}
    \resizebox{\textwidth}{!}{
      \begin{tabular}{lccc|ccc|ccc|ccc|ccc}
      \toprule
          \multicolumn{1}{c}{}
          & \multicolumn{3}{c}{\makecell{\textbf{LIBERO-Spatial}}}
          & \multicolumn{3}{c}{\makecell{\textbf{LIBERO-Object}}}
          & \multicolumn{3}{c}{\makecell{\textbf{LIBERO-Goal}}}
          & \multicolumn{3}{c}{\makecell{\textbf{LIBERO-Long}}}
          & \multicolumn{3}{c}{\makecell{\textbf{Overall}}} \\
          \cmidrule(lr){2-4}\cmidrule(lr){5-7}\cmidrule(lr){8-10}\cmidrule(lr){11-13}\cmidrule(lr){14-16}
          \multirow{2}{*}{\makecell{\textbf{Victim VLA}}}
          & \makecell{\textbf{Base}\\ \textbf{CV} $\downarrow$}
          & \makecell{\textbf{Attack}\\ \textbf{CV} $\uparrow$}
          & \makecell{\textbf{CVI}\\ $\Delta \text{CV}$ $\uparrow$}
          & \makecell{\textbf{Base}\\ \textbf{CV} $\downarrow$}
          & \makecell{\textbf{Attack}\\ \textbf{CV} $\uparrow$}
          & \makecell{\textbf{CVI}\\ $\Delta \text{CV}$ $\uparrow$}
          & \makecell{\textbf{Base}\\ \textbf{CV} $\downarrow$}
          & \makecell{\textbf{Attack}\\ \textbf{CV} $\uparrow$}
          & \makecell{\textbf{CVI}\\ $\Delta \text{CV}$ $\uparrow$}
          & \makecell{\textbf{Base}\\ \textbf{CV} $\downarrow$}
          & \makecell{\textbf{Attack}\\ \textbf{CV} $\uparrow$}
          & \makecell{\textbf{CVI}\\ $\Delta \text{CV}$ $\uparrow$}
          & \makecell{\textbf{Base}\\ \textbf{CV} $\downarrow$}
          & \makecell{\textbf{Attack}\\ \textbf{CV} $\uparrow$}
          & \makecell{\textbf{CVI}\\ $\Delta \text{CV}$ $\uparrow$} \\
          \midrule
  
          \textsc{$\pi_0$-LIBERO}~\cite{intelligence2025pi_}
          & 550.7 & 326.2 & 0.59
          & 711.5 & 838.3 & 1.18
          & 309.6 & 624.5 & 2.02
          & 1039.6 & 1269.1 & 1.22
          & 652.9 & 764.5 & 1.17 \\

          \textsc{$\pi_{0.5}$}~\cite{intelligence2025pi_}
          & 570.9 & 549.8 & 0.96
          & 681.4 & 699.0 & 1.03
          & 260.3 & 439.1 & 1.69
          & 863.9 & 1336.2 & 1.55
          & 594.1 & 756.0 & 1.27 \\

          \textsc{GR00T-N1.5}~\cite{bjorck2025gr00t}
          & 599.9 & 702.7 & 1.17
          & 644.1 & 595.7 & 0.92
          & 258.9 & 862.3 & 3.33
          & 918.8 & 778.1 & 0.85
          & 605.4 & 734.7 & 1.21 \\

          \textsc{X-VLA}~\cite{zheng2025x}
          & 838.3 & 1356.3 & 1.62
          & 828.3 & 725.7 & 0.88
          & 347.1 & 885.0 & 2.55
          & 1145.3 & 1171.0 & 1.02
          & 789.8 & 1034.5 & 1.31 \\

          \midrule

          \textsc{InternVLA-M1}~\cite{chen2025internvla}
          & 475.9 & 130.3 & 0.27
          & 639.3 & 493.0 & 0.77
          & 232.9 & 495.3 & 2.13
          & 681.9 & 1550.3 & 2.27
          & 507.5 & 667.2 & 1.31 \\

          \textsc{DeepThinkVLA}~\cite{yin2025deepthinkvla}
          & 572.4 & 759.7 & 1.33
          & 607.9 & 729.0 & 1.20
          & 220.2 & 915.7 & 4.16
          & 827.1 & 1509.7 & 1.83
          & 556.9 & 978.5 & 1.76 \\

          \midrule
          \textbf{Average}
          & \textbf{601.4} & \textbf{637.5} & \textbf{1.06}
          & \textbf{685.4} & \textbf{680.1} & \textbf{0.99}
          & \textbf{271.5} & \textbf{703.7} & \textbf{2.59}
          & \textbf{912.8} & \textbf{1269.1} & \textbf{1.39}
          & \textbf{617.8} & \textbf{822.6} & \textbf{1.33} \\
          \bottomrule
      \end{tabular}
      }
      \end{center}
      \vspace{-5pt}
    \caption{\textbf{LIBERO category-level constraint violation results across victim VLA models.}
  Columns are grouped by the four LIBERO suites (\textbf{Spatial}, \textbf{Object}, \textbf{Goal}, \textbf{Long}) and \textbf{Overall}.
  We report \textbf{Base CV} (average constraint violations per episode without attack), \textbf{Attack CV} (average constraint violations per episode under attack),
  and \textbf{CVI} (constraint violation inflation) $\Delta \text{CV}=\text{CV}_{\text{attack}} / \text{CV}_{\text{base}}$.
  The \textbf{Average} row reports the mean over victim models for each column.}
    \label{tab:libero_victims_by_category_constraint_violation_cvi}
    \vspace{-12pt}
  \end{table*}

\subsection{Experimental Setup}
\noindent \textbf{Benchmark.} We leverage LIBERO~\cite{liu2023libero}, a popular language-conditioned robot manipulation benchmark covering diverse household tasks, \textit{e.g.}, pick-and-place, object rearrangement, drawer/door opening, and multi-step long-horizon manipulation, to evaluate the effectiveness of our adversarial attacks on VLA policies.

\noindent \textbf{Target VLA models.}
We select target VLA policies that are \emph{strong and well-established} on LIBERO, each achieving over 90\% task success under standard evaluation, so our attacks are tested against robust, high-performing baselines rather than weak policies. Concretely, our target set includes widely used open-source VLAs: $\pi_{0}$~\cite{intelligence2025pi_}, $\pi_{0.5}$~\cite{intelligence2025pi_}, X-VLA~\cite{zheng2025x}, and GR00T-N1.5~\cite{bjorck2025gr00t}; and deliberative/reasoning-centric VLAs that expose intermediate reasoning signals (e.g., chain-of-thought or structured traces), including DeepThinkVLA~\cite{yin2025deepthinkvla} and InternVLA-M1~\cite{chen2025internvla}.

\noindent \textbf{Metrics.} We report two groups of metrics for agentic attack performance. \emph{Objective} metrics include: (i) \textbf{ASR} (attack success ratio) for \emph{task failure}, measured as the drop in task success rate from the baseline VLA to the perturbed (attacked) VLA; (ii) \textbf{AIR} (action inflation ratio), measured as how many times longer the action sequence becomes under attack compared to the baseline; and (iii) \textbf{CVI} (constraint violation inflation), measured as how many times more constraint violations occur under attack compared to the baseline.
\emph{Stealth} metrics include: (i) \textbf{Tool Calls}, the average number of tool calls per episode and (ii) \textbf{Char Edits}, the character-level edit distance per episode (average number of modified characters).


\noindent \textbf{Cold-start initialization.}
Figure~\ref{fig:training} shows our two-stage training procedure. In the \textbf{cold-start} stage, we bootstrap the attack agent with a small set of tool-use trajectories through supervised fine-tuning (SFT), enabling reliable execution of the \texttt{FIND}$\rightarrow$\texttt{APPLY} protocol and effective text perturbation generation. In the second stage, we optimize the initialized agent with \textbf{GRPO} in a black-box setting, using objective-specific rollout rewards, optionally with a stealth penalty, and grouped rollouts per scenario to compute advantages while updating only the attacker’s LoRA parameters. We instantiate the attacker with Qwen2.5-3B-Instruct~\cite{qwen2025qwen25technicalreport} and train three separate agents, one per attack objective. For each objective, we build the cold-start dataset with GPT-5-mini~\cite{openai_gpt5mini_model_2026} under the same scenario interface, then fine-tune the Qwen-based attacker before GRPO to improve early exploration and stabilize training.

\noindent \textbf{Training setting.}
We split LIBERO evaluation tasks into disjoint train/test sets for red-team learning. We train the attack agent against $\pi_{0.5}$ on the first $7$ evaluation tasks from each of LIBERO’s four suites (\emph{Goal}, \emph{Object}, \emph{Spatial}, and \emph{Long-horizon}), with $8$ episodes per task, and test transfer on the remaining $3$ held-out tasks per suite on $\pi_{0.5}$ and other target VLA models over $5$ episodes per task. During an episode, the attacker is constrained to at most $200$ character-level edits (Levenshtein budget) and can invoke the tool-chain up to $4$ times via a \texttt{FIND}$\rightarrow$\texttt{APPLY} protocol.

\subsection{Attack Results by Across Objectives and Task Suites}

\noindent \textbf{Results by attack objective.} We evaluate three attack objectives: \emph{Task Failure} (Table~\ref{tab:libero_victims_by_category_task_failure}), \emph{Action Inflation} (Table~\ref{tab:libero_victims_by_category_action_length_air}), and \emph{Constraint Violation} (Table~\ref{tab:libero_victims_by_category_constraint_violation_cvi}). For each objective, we report the aligned metric: \textbf{ASR} for \emph{Task Failure}, \textbf{AIR} for \emph{Action Inflation}, and \textbf{CVI} for \emph{Constraint Violation}, along with baseline and target rollout statistics for context. Across diverse target VLAs, our attacker induces consistent objective-driven degradation, reducing task success by 20.6\%, increasing action-sequence length by 55\%, and raising constraint violations by 33\% on average, demonstrating strong transferability across models with and without explicit reasoning. 

\noindent \textbf{Attack performance across task suites.} Attack efficacy also varies with task structure: perception- and grounding-heavy suites such as LIBERO \emph{Spatial} and \emph{Object} are generally less amplified for task failure and constraint violation, and perturbed instructions can occasionally trigger safer or more efficient replanning; by contrast, planning-heavy suites such as LIBERO \emph{Goal} and \emph{Long} are more vulnerable, especially for inducing failures and physical constraint violations. \emph{Action Inflation} remains effective across all suites, with particularly strong gains on shorter-horizon tasks while still transferring to long-horizon settings.


\begin{table*}[t]
  \vspace{0.4em}
  \centering
  \resizebox{0.9\textwidth}{!}{
    \begin{tabular}{lcc|cc|cc|cc|cc}
    \toprule
        \multicolumn{1}{c}{}
        & \multicolumn{2}{c}{\makecell{\textbf{Spatial}}}
        & \multicolumn{2}{c}{\makecell{\textbf{Object}}}
        & \multicolumn{2}{c}{\makecell{\textbf{Goal}}}
        & \multicolumn{2}{c}{\makecell{\textbf{Long}}}
        & \multicolumn{2}{c}{\makecell{\textbf{Overall}}} \\
        \cmidrule(lr){2-3}\cmidrule(lr){4-5}\cmidrule(lr){6-7}\cmidrule(lr){8-9}\cmidrule(lr){10-11}
        \makecell{\textbf{Objective}}
        & \makecell{\textbf{Tool}\\ \textbf{Calls} $\downarrow$}
        & \makecell{\textbf{Char}\\ \textbf{Edits} $\downarrow$}
        & \makecell{\textbf{Tool}\\ \textbf{Calls} $\downarrow$}
        & \makecell{\textbf{Char}\\ \textbf{Edits} $\downarrow$}
        & \makecell{\textbf{Tool}\\ \textbf{Calls} $\downarrow$}
        & \makecell{\textbf{Char}\\ \textbf{Edits} $\downarrow$}
        & \makecell{\textbf{Tool}\\ \textbf{Calls} $\downarrow$}
        & \makecell{\textbf{Char}\\ \textbf{Edits} $\downarrow$}
        & \makecell{\textbf{Tool}\\ \textbf{Calls} $\downarrow$}
        & \makecell{\textbf{Char}\\ \textbf{Edits} $\downarrow$} \\
        \midrule
        \textsc{Task Failure}
        & 2.97 & 13.4
        & 3.34 & 13.2
        & 2.80 & 10.3
        & 2.97 & 15.1
        & 3.02 & 13.0 \\
        \textsc{Action Inflation}
        & 2.98 & 126.7
        & 3.26 & 114.0
        & 3.36 & 130.3
        & 3.05 & 117.3
        & 3.16 & 122.1 \\
        \textsc{Constraint Violation}
        & 3.34 & 89.3
        & 2.34 & 65.0
        & 1.67 & 50.0
        & 2.34 & 51.7
        & 2.42 & 64.0 \\
        \bottomrule
    \end{tabular}
  }
  \vspace{-3pt}
  \caption{\textbf{Average tool usage by objective.}
Mean \textbf{Tool Calls} and \textbf{Char Edits} per episode for each LIBERO suite and averaged over victim VLA models.}
  \label{tab:libero_tool_usage_by_objective}
  \vspace{-15pt}
\end{table*}
\begin{table}[t]
    \vspace{0.4em}
    \centering
    \resizebox{\columnwidth}{!}{
    \begin{tabular}{l|cc|ccc}
      \toprule
      \makecell{\textbf{Models}}
      & \multicolumn{2}{c}{\textbf{Stealthy}}
      & \multicolumn{3}{c}{\textbf{Objective}} \\
      \cmidrule(lr){2-3} \cmidrule(lr){4-6}
      & \makecell{\textbf{Tool}\\\textbf{Calls} $\downarrow$} &
      \makecell{\textbf{Char}\\\textbf{Edit} $\downarrow$} &
      \makecell{\textbf{ASR} $\uparrow$} & \makecell{\textbf{AIR}\\ $\Delta|\mathbf{a}|$ $\uparrow$} & \makecell{\textbf{CVI}\\ $\Delta \text{CV}$ $\uparrow$} \\
      \midrule
      \makecell{\textsc{GPT-5 mini}} & 3.93 & 168.8 & 14.5 & 1.37 & 1.25 \\
      \makecell{\textsc{\ours{}}} & 3.10 & 76.46 & 16.7 & 1.38 & 1.27 \\
      \bottomrule
    \end{tabular}}
    \caption{\textbf{Objective-level summary of attack effectiveness and stealth.}
We compare a frozen \textsc{GPT-5 mini} attacker (same tool-calling interface) against \ours{} on stealthy and Objective metrics.
\ours{} achieves comparable or better objective performance while using substantially fewer tool calls and character edits, indicating more efficient, high-leverage perturbations learned via RL.}
\vspace{-10pt}
    \label{tab:objective_level_summary}
  \end{table}
\begin{table}[t]
    \vspace{0.4em}
    \centering
    \resizebox{\columnwidth}{!}{
    \begin{tabular}{l|cc|ccc}
      \toprule
      \makecell{\textbf{Models}}
      & \multicolumn{2}{c}{\textbf{Stealthy}}
      & \multicolumn{3}{c}{\textbf{Objective}} \\
      \cmidrule(lr){2-3} \cmidrule(lr){4-6}
      & \makecell{\textbf{Tool}\\\textbf{Calls} $\downarrow$} &
      \makecell{\textbf{Char}\\\textbf{Edit} $\downarrow$}
      & \makecell{\textbf{Base}\\ \textbf{TER} $\uparrow$}
      & \makecell{\textbf{Attack}\\ \textbf{TER} $\downarrow$}
      & \makecell{\textbf{ASR} $\uparrow$} \\
      \midrule
      \makecell{\textsc{GRPO Only}} & 2.76 &  11.78 &  96.7 & 88.0 & 8.7 \\
    \makecell{\textsc{SFT + GRPO}} & 3.10 & 76.46 & 96.7  & 80.0 & 16.7 \\
      \bottomrule
    \end{tabular}}
\caption{\textbf{Necessity of cold-start.} We ablate agent training with cold-start data (\textsc{SFT}+\textsc{GRPO}) versus without cold-start data (\textsc{GRPO} only) under the \emph{Task Failure} objective. Cold-start data is crucial for stabilizing RL training and preventing policy degradation.}
\vspace{-12pt}
    \label{tab:cold-start}
  \end{table}

\subsection{Tool-Use Strategy and Attack Efficiency}
\label{sec:exp:tool_call}


Table~\ref{tab:libero_tool_usage_by_objective} reports stealth metrics, including the average number of tool calls and character edits per episode. Clear differences emerge across objectives under the same edit budget: \emph{Action Inflation} uses tools most frequently and makes the most edits, whereas \emph{Task Failure} uses the fewest edits, only 10.6\% of the budget. \emph{Constraint Violation} also requires fewer tool calls on average, by 23.4\%. These objective-dependent patterns motivate a closer examination of tool-use evolution during training.
Inspection of policy evolution reveals a clear shift from exploratory tool use in cold-start rollouts to more selective strategies after GRPO training. In particular, \emph{char-level} edits contribute only marginally to successful attacks, as they are strongly constrained by the tool-call and edit budgets. By contrast, \emph{token-level} perturbations dominate after training, especially for \emph{Task Failure}, because they achieve higher reward with fewer tool calls and edits. Overall, the learned agents adapt their tool-use strategies to each objective.

\subsection{Comparison to Baselines and Training Analysis}

\noindent \textbf{\ours ~v.s. GPT-based attacker:}
We compare \ours{} with a GPT-based attacker that uses the same tool-calling interface as a strong frozen baseline (Table~\ref{tab:objective_level_summary}). Relative to this baseline, \ours{} achieves consistent gains on objective-aligned metrics, with about a 2\% absolute improvement on average across objectives, indicating measurable policy learning beyond prompt engineering alone. More importantly, \ours{} is substantially more efficient and stealthy, requiring 21.1\% fewer tool calls and 54.7\% fewer character edits to achieve comparable or better attack outcomes. This suggests that RL training not only improves attack effectiveness, but also learns higher-leverage perturbation strategies that reduce unnecessary interactions and superficial instruction changes, which is particularly desirable in practical black-box attacks where excessive tool usage or large edits are easier to detect.

\noindent \textbf{Effect of cold-start training:}
We further study the role of the cold-start stage in our training pipeline. Table~\ref{tab:cold-start} reports an ablation comparing training with and without cold-start initialization. Cold-start data is important for stabilizing RL and preventing policy degradation: when SFT initialization is removed and the agent is trained with \textsc{GRPO} alone under the \emph{Task Failure} objective, the resulting policy does not improve reliably through RL. Although it uses fewer tool calls and makes fewer character edits, its attack effectiveness drops, with \textbf{ASR} decreasing by 5.8\% relative to the SFT+\textsc{GRPO} variant. This suggests that, without cold-start supervision, the agent struggles to discover effective tool-calling behaviors and may instead over-optimize for lower-cost actions that do not translate into stronger attacks.

\subsection{Discussion}


\noindent \textbf{Adaptation across objectives and task regimes.}
%
%
Our pipeline generalizes across objectives, target VLAs, and task suites by learning objective-specific tool-use and perturbation strategies. Across six strong VLA policies, the trained agents consistently induce objective-aligned degradation: lower task success, longer action sequences, and more constraint violations, showing strong transferability to models with and without explicit reasoning. By generating feasible instruction-level attacks for both direct and indirect objectives, the learned policies reduce reliance on hand-designed heuristics and improve scalability across tasks.


\noindent \textbf{Transition from effectiveness to efficiency.}
%
%
Overall, the policy appears to be learned in two stages: first, identifying a feasible attack pattern, then refining it into smaller, higher-leverage perturbations. 
Early on, the agent relies on broader prompt-level perturbations, which are costly but useful for discovering feasible strategies across tasks and models. As training continues, it shifts toward more selective token-level edits that preserve attack success while reducing tool usage and edit cost. This trend is most evident for \emph{Task Failure}, where effectiveness is maintained despite fewer tool calls and character edits. 


\noindent \textbf{Cold-start stabilizes tool discovery.}
Ablation results show that \textsc{GRPO} alone does not reliably learn strong attacks from scratch. Without cold-start supervision, the agent tends to converge to low-cost but ineffective behaviors, \textit{i.e.}, fewer tool calls and edits without improved attack success. This suggests that the challenge lies not only in reward optimization, but also in discovering valid tool-use patterns and feasible perturbation templates in a large discrete action space. Cold-start provides this initial scaffold, after which \textsc{GRPO} refines the policy into more effective and lower-cost attacks. It is therefore a key component for stable agentic attack learning.

\section{Conclusion}
\label{sec:conclusion}

We presented \ours{}, an agent-centric black-box framework for stealthy, instruction-only attacks on vision-language-action (VLA) policies. With a multi-turn ReAct attacker trained by \textsc{GRPO}, \ours{} performs bounded \texttt{FIND}$\rightarrow$\texttt{APPLY} edits to generate objective-driven perturbations without gradients or target-specific redesign. Across \emph{task failure}, \emph{action inflation}, and \emph{constraint violation}, it induces targeted degradation on LIBERO while using fewer tool calls and character edits than strong GPT-based baselines, suggesting higher-leverage and less detectable attacks. These results highlight learned attacker models as a practical and scalable tool for red-teaming robotic foundation models.
Our study is currently limited to text-only perturbations in simulation, leaving multimodal and real-world physical attack surfaces unexplored. Future work will extend attacks to the victim’s reasoning process, including intermediate plans and tool-selection logic during execution, and to multimodal settings that jointly perturb language and perception to probe cross-modal vulnerabilities under realistic deployment constraints.

\footnotesize{
\bibliographystyle{ieeetr}
\bibliography{custom}
}

\end{document}